\documentclass{article}

\usepackage{microtype}
\usepackage{graphicx}
\usepackage{subcaption}
\usepackage{booktabs} 

\usepackage{hyperref}

\usepackage[preprint]{icml2026}

\usepackage{amsmath}
\usepackage{amssymb}
\usepackage{mathtools}
\usepackage{amsthm}
\usepackage{enumitem}
\usepackage{tcolorbox}
\usepackage{algorithm}
\usepackage{algorithmic}
\usepackage{xcolor}

\newcommand{\sysname}{\textsc{ScholarGym}}

\usepackage[capitalize,noabbrev]{cleveref}

\theoremstyle{plain}

\theoremstyle{definition}

\theoremstyle{remark}

\usepackage[textsize=tiny]{todonotes}

\icmltitlerunning{ScholarGym: Benchmarking LLM Capabilities in Deep Research}

\begin{document}

\twocolumn[
  \icmltitle{ScholarGym: Benchmarking Large Language Model Capabilities in the Information-Gathering Stage of Deep Research}



  \icmlsetsymbol{equal}{*}

  \begin{icmlauthorlist}
    \icmlauthor{Hao Shen}{same}
    \icmlauthor{Hang Yang}{same}
    \icmlauthor{Zhouhong Gu}{same}
    \icmlauthor{Weili Han}{same}
  \end{icmlauthorlist}

  \icmlaffiliation{same}{Fudan University}

  \icmlcorrespondingauthor{Hao Shen}{hshen22@m.fudan.edu.cn}
  \icmlcorrespondingauthor{Hang Yang}{hangyang22@m.fudan.edu.cn}
  \icmlcorrespondingauthor{Zhouhong Gu}{zhgu22@m.fudan.edu.cn}
  \icmlcorrespondingauthor{Weili Han}{wlhan@fudan.edu.cn}

  \icmlkeywords{Machine Learning, ICML}

  \vskip 0.3in
]



\printAffiliationsAndNotice{}  

\begin{abstract}
  Large language models have advanced from single-turn question answering to deep research systems that iteratively decompose research questions, invoke retrieval tools, and synthesize information across multiple rounds. Evaluating such systems typically involves scoring their final research reports holistically, but this end-to-end paradigm tightly couples the language model's decision-making, workflow design, and environmental feedback, precluding decomposable analysis of individual components.

  We introduce \sysname, an evaluation environment that isolates the information-gathering stage of deep research on academic literature. Under a unified workflow, \sysname decomposes the research process into three explicit stages---Query Planning, Tool Invocation, and Relevance Assessment---and evaluates each against 2,536 expert-annotated queries over a static corpus of 570K papers with deterministic retrieval. Systematic experiments reveal that iterative query decomposition yields 2.9--3.3$\times$ F1 gains over single-query retrieval, models with extended thinking trade recall for precision, and Query Planning quality together with Relevance Assessment constitute dual bottlenecks that separate proprietary from open-source model performance.
\end{abstract}

\section{Introduction}
\label{sec:intro}

In recent years, large language models (LLMs) have evolved from single-turn question answering to deep research systems capable of long-horizon, iterative information gathering and synthesis~\cite{li2025search,wu2025webdancer,zhang2025evolvesearch}. In academic literature retrieval, such systems progressively decompose a research question into actionable subqueries, repeatedly invoke retrieval tools, and dynamically adjust their search and filtering strategies through multi-round feedback, ultimately integrating relevant findings into structured research reports or survey-style analyses~\cite{wang2024autosurvey,bao2025surveygen,he2025pasa}. Commercial systems operating in this paradigm---including OpenAI Deep Research, Gemini Deep Research, and Kimi Agent~\cite{openaideepresearch,geminideepresearch,kimiagent}---have demonstrated substantial practical value.

As deep research systems advance rapidly, numerous efforts have attempted to evaluate this class of tasks, typically by running a complete deep research pipeline and scoring the final research report holistically~\cite{gou2025mind2web,du2025deepresearch,li2025reportbench}. While such evaluations provide useful reference points for overall system performance under particular configurations, their principal limitation is that multiple critical factors---the language model's decision-making, tool interface design, workflow orchestration, and environmental feedback---remain tightly coupled throughout evaluation. These factors mutually influence and constrain one another, making it difficult to attribute observed outcomes to specific components. Under this setup, evaluation results can only indicate \emph{whether} the system works as a whole, but cannot answer \emph{which} components are effective and \emph{which} decisions truly determine research quality.

A principled evaluation framework for deep research should satisfy several key requirements. First, evaluation should center on the language model's decisions and actions, enabling the contribution of each component to be explicitly disentangled and analyzed, rather than relying solely on end-to-end judgment. Second, since final reports vary widely in structure and style, evaluation should target the upstream behaviors that most determine report quality---information gathering, coverage, and filtering---that have a decisive impact on research quality. Third, to support reproducible and comparable analysis, the evaluation environment should eliminate the non-determinism introduced by live web APIs and open-ended search, characterizing model and component interactions under controlled conditions.

We introduce \sysname, an evaluation environment that targets the information-gathering stage of deep research on academic literature. \sysname fixes a unified workflow and decomposes it into three stages---Query Planning, Tool Invocation, and Relevance Assessment---with the language model as the sole decision-maker at each stage. All retrieval operates over a static corpus of 570K papers with deterministic ranking, enabling systematic comparison of model behavior during information gathering. We construct 2,536 queries with expert-annotated ground truth and equip the workflow with a memory mechanism that maintains reasoning state across iterations, enabling rigorous, reproducible evaluation of long-horizon academic retrieval.

Through systematic evaluation of mainstream LLMs in the  \sysname environment, we reveal key behavioral patterns and performance bottlenecks in deep research. After five iterations, GPT-5.2 achieves the best F1 (0.447) by balancing recall (0.837) and precision (0.305), while Gemini3-Pro attains the highest recall (0.950) at lower precision. Models with ``extended thinking'' capabilities tend to sacrifice recall for higher precision. Decomposing complex research questions into subqueries and searching iteratively yields 2.9--3.3$\times$ F1 improvements over single-query retrieval, confirming the value of iterative planning. We further identify Query Planning quality and Relevance Assessment as dual bottlenecks constraining open-source models: proprietary models score 43\% higher on Avg.Distance, reflecting a marked advantage in query formulation.

\section{Related Work}

\subsection{Deep Research Evaluation}
Benchmarks for deep research have evolved from factoid QA to multi-step information synthesis. GAIA~\cite{mialon2023gaia} and BrowseComp~\cite{wei2025browsecomp} evaluate tool use and web navigation but focus on intermediate reasoning rather than comprehensive outputs. DeepResearch Bench~\cite{du2025deepresearch,li2026deepresearch}, Rigorous Bench~\cite{yao2025rigorous}, DEER~\cite{han2025deer}, and FINDER~\cite{zhang2025far} shift toward holistic evaluation of report quality using multidimensional rubrics.

A shared limitation is reliance on live APIs. \citet{chen2025xbench} observe that temporal drift in web content creates a ``reproducibility gap'' obscuring algorithmic progress. \citet{zhang2025far} report that evaluation failures frequently stem from anti-scraping mechanisms and URL timeouts rather than model deficiencies. LiveResearchBench~\cite{wang2025liveresearchbench} acknowledges that the absence of fixed ground truth renders verification costly and noisy. These observations motivate controlled evaluation that decouples workflow capabilities from environmental variance.

\subsection{Academic Literature Retrieval}
Retrieving scientific literature introduces additional challenges: citation networks are dense, terminology is specialized, and relevance judgments require domain expertise. PaSa~\cite{he2025pasa} proposes an autonomous agent for paper search but relies on Google Search APIs, inheriting the non-determinism of general web agents. CiteME~\cite{press2024citeme} shows that API inconsistencies and broken links cause state-of-the-art models to fail on citation tasks due to environmental friction alone.

Static benchmarks offer a partial remedy. LitSearch~\cite{ajith2024litsearch} constructs a fixed corpus from ACL and ICLR papers and finds that dense retrievers outperform commercial search engines on deep understanding queries. SciNetBench~\cite{shao2025scinetbench} uses an OpenAlex snapshot for relation-aware retrieval. HiSciBench~\cite{zhang2025hiscibench} highlights that literature review generation suffers from factuality gaps exacerbated by dynamic sources. Our work extends this direction by coupling a static corpus with an iterative workflow, enabling reproducible assessment of multi-turn planning and selection.

\subsection{Deep Research Systems}
Current deep research systems fall into two categories. \textbf{Workflow-based} systems decompose queries through explicit planning stages. Alita~\cite{qiu2025alita} enables scalable agentic reasoning with minimal predefinition. OWL~\cite{hu2025owl} optimizes multi-agent coordination for task automation. WisPaper~\cite{ju2025wispaper} targets academic search via structured query expansion. These frameworks expose workflow logic but depend on live web APIs, causing execution paths to diverge across runs.

\textbf{RL-based} systems train agents end-to-end on search trajectories. DeepResearcher~\cite{zheng2025deepresearcher} applies reinforcement learning in real-world environments. Search-R1~\cite{jin2025search} and SimpleTIR~\cite{xue2025simpletir} train on multi-turn tool interactions. DeepDive~\cite{lu2025deepdive} integrates knowledge graphs with multi-turn RL. \citet{gao2025beyond} and \citet{li2025nested} scale asynchronous RL for long-horizon search. Tongyi DeepResearch~\cite{team2025tongyi} validates algorithms on a Wikipedia snapshot to address non-stationarity—a ``wind tunnel'' strategy that parallels our motivation. Step-DeepResearch~\cite{hu2025step} synthesizes atomic capabilities to bypass unstable API feedback.

Both paradigms face the same obstacle: live web APIs inject distributional drift that confounds learning signals~\cite{xue2025simpletir}. \sysname provides a static environment where workflow design and RL training can proceed without environmental noise, enabling controlled comparison across systems.

\section{ScholarGym}
\label{sec:method}

\begin{figure*}[t]
  \includegraphics[width=1\linewidth]{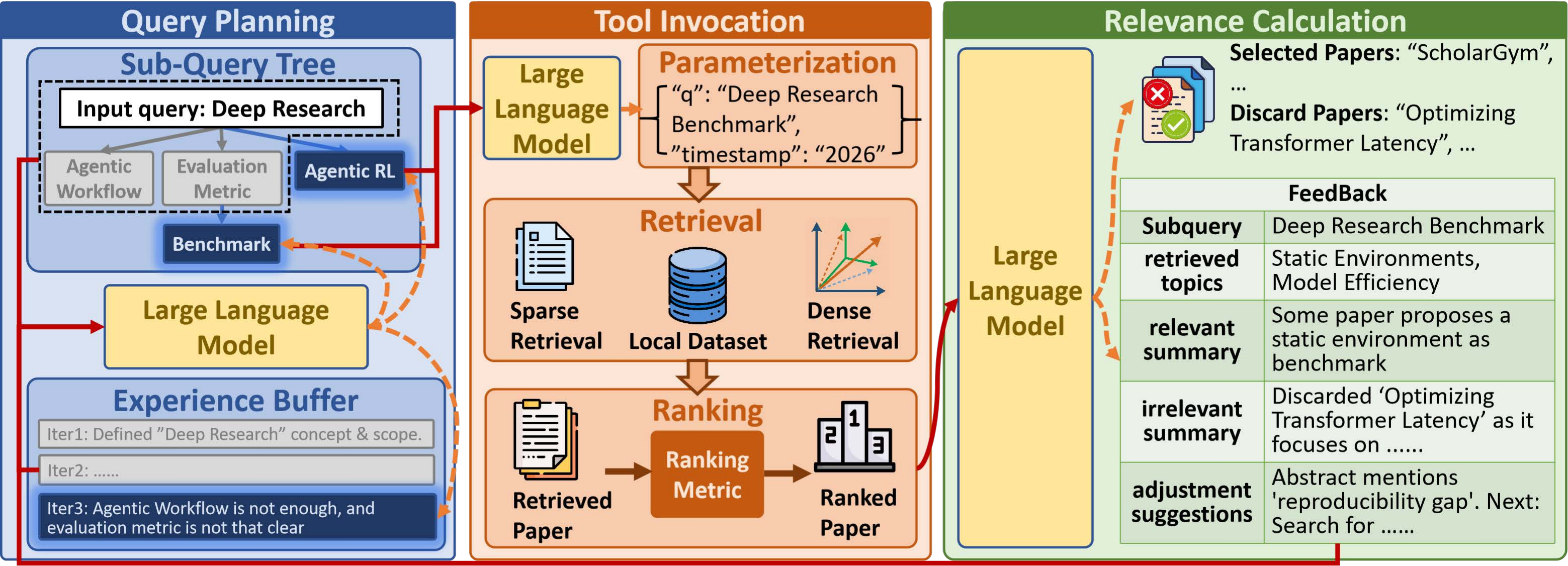}
  \caption{Overview of \sysname. Each iteration comprises three stages connected by directed information flow: solid red arrows indicate stage inputs, and dashed orange arrows indicate outputs. \textbf{Query Planning} receives the subquery tree $\mathcal{M}_{t-1}$, experience buffer $\mathcal{B}_{t-1}$, and feedback $\mathcal{O}_{t-1}$ from the previous iteration, then outputs newly generated child nodes and an updated experience buffer. The resulting subqueries are passed to \textbf{Tool Invocation}, which parameterizes retrieval calls and executes them against the corpus; retrieved candidates undergo relevance-based ranking before selection. \textbf{Relevance Assessment} evaluates ranked papers and produces feedback to guide the next iteration.}
  \label{fig:framework}
  \vspace{-3mm}
\end{figure*}

We present \sysname, a simulation environment for reproducible evaluation of deep research workflows. The environment consists of two components: (1)~an iterative deep research workflow with closed-loop feedback, and (2)~a benchmark suite containing expert-annotated queries and a unified paper corpus. Figure~\ref{fig:framework} illustrates the architecture; Table~\ref{tab:notation} summarizes notation.

\begin{table}[t]
\caption{Notation. Subscript $t$ denotes the iteration index.}
\label{tab:notation}
\centering
\small
\begin{tabular}{@{}cl@{}}
\toprule
\textbf{Symbol} & \textbf{Description} \\
\midrule
$q$, $d$, $\mathcal{D}$ & Research query, date constraint, paper corpus \\
$\mathcal{G}$ & Ground-truth paper set \\
$\mathcal{M}_t$ & Subquery tree at iteration $t$ \\
$\mathcal{B}_t$ & Experience buffer at iteration $t$ \\
$\mathcal{P}_t$ & Search plan at iteration $t$ \\
$\mathcal{Q}_t$ & Parameterized tool calls at iteration $t$ \\
$\mathcal{C}_i$ & Candidate set from $i$-th tool call \\
$\mathcal{S}_t$, $\mathcal{O}_t$ & Selected papers, feedback at iteration $t$ \\
\bottomrule
\end{tabular}
\vskip -0.1in
\end{table}

\subsection{Deep Research Workflow}
\label{sec:workflow}

Given a research query $q$, the workflow decomposes it into \emph{subqueries}---specialized search terms targeting specific aspects of $q$---and iteratively expands this set based on retrieval outcomes. Each iteration comprises three stages: (1)~\textbf{query planning} analyzes search progress and proposes new subqueries; (2)~\textbf{tool invocation} generates and executes parameterized retrieval calls; and (3)~\textbf{relevance assessment} evaluates candidate papers and provides feedback $\mathcal{O}_t$ to guide the next iteration. The process terminates when no new subqueries are proposed or after $T$ iterations. Algorithm~\ref{alg:workflow} presents the procedure.

\begin{algorithm}[t]
\caption{Deep Research Workflow}
\label{alg:workflow}
\small
\begin{algorithmic}[1]
\STATE \textbf{Input:} Query $q$, date constraint $d$, corpus $\mathcal{D}$, max iterations $T$
\STATE \textbf{Output:} Selected paper set $\mathcal{S}$
\STATE \textcolor{gray}{// Initialization}
\STATE $\mathcal{M}_0 \leftarrow \{q\}$, $\mathcal{B}_0 \leftarrow \emptyset$, $\mathcal{O}_0 \leftarrow \emptyset$, $\mathcal{S} \leftarrow \emptyset$
\FOR{$t = 1$ to $T$}
    \STATE \textcolor{gray}{// Stage 1: Query Planning -- analyze progress and propose subqueries}
    \STATE $(\mathcal{P}_t, \mathcal{M}_t, \mathcal{B}_t) \leftarrow \textsc{Plan}(q, \mathcal{M}_{t-1}, \mathcal{B}_{t-1}, \mathcal{O}_{t-1})$
    \IF{$\mathcal{P}_t = \emptyset$}
        \STATE \textbf{return} $\mathcal{S}$ \hfill \textcolor{gray}{// no new subqueries, search converged}
    \ENDIF
    \STATE \textcolor{gray}{// Stage 2: Tool Invocation -- generate and execute tool calls}
    \STATE $\mathcal{Q}_t \leftarrow \textsc{Parameterize}(\mathcal{P}_t)$
    \STATE $\{\mathcal{C}_i\}_{i=1}^{|\mathcal{Q}_t|} \leftarrow \textsc{Execute}(\mathcal{Q}_t, d, \mathcal{D})$
    \STATE \textcolor{gray}{// Stage 3: Relevance Assessment -- evaluate and select papers}
    \STATE $(\mathcal{S}_t, \mathcal{O}_t) \leftarrow \textsc{Assess}(\{\mathcal{C}_i\}, \mathcal{P}_t)$
    \STATE $\mathcal{S} \leftarrow \mathcal{S} \cup \mathcal{S}_t$
\ENDFOR
\STATE \textbf{return} $\mathcal{S}$
\end{algorithmic}
\end{algorithm}

\paragraph{Query Planning.}
At iteration $t$, this stage receives the subquery tree $\mathcal{M}_{t-1}$, experience buffer $\mathcal{B}_{t-1}$, and feedback $\mathcal{O}_{t-1}$ from the previous iteration. It identifies underexplored aspects of the research question and produces a plan $\mathcal{P}_t$ specifying new subqueries. The subquery tree is rooted at the original query $q$, where each node represents a subquery derived from its parent through semantic refinement.

New subqueries are proposed through three actions: \textbf{Derive} creates a more specific subquery from an existing one (e.g., ``transformer efficiency'' $\rightarrow$ ``sparse attention''); \textbf{Expand} creates a sibling subquery exploring a different aspect; \textbf{Continue} requests additional results for an existing subquery. The plan $\mathcal{P}_t$ contains the proposed subqueries with semantic descriptions and relevance criteria to guide subsequent assessment.

\paragraph{Tool Invocation.}
This stage receives plan $\mathcal{P}_t$ and generates parameterized tool calls $\mathcal{Q}_t$. For each subquery in $\mathcal{P}_t$, it formulates a structured call $(q_i, k_i, d_i)$ specifying the query string $q_i$, number of results $k_i$, and date constraint $d_i$.

Tool invocation comprises two sub-stages following standard retrieval practice. \textbf{Retrieval} executes calls $\mathcal{Q}_t$ against corpus $\mathcal{D}$ using either sparse or dense methods, returning an initial candidate set for each query. Sparse methods excel at exact term matching, while dense methods capture semantic relationships beyond lexical overlap. \textbf{Ranking} reorders candidates within each set by relevance scores derived from the underlying retrieval method. The ranked sets $\{\mathcal{C}_i\}$ are passed to relevance assessment, where each $\mathcal{C}_i$ contains up to $k_i$ papers with metadata (title, abstract, publication date, identifier). Implementation details appear in Appendix~\ref{sec:tool_details}.

\paragraph{Relevance Assessment.}
This stage evaluates each candidate paper against the research objective using the criteria in $\mathcal{P}_t$. We consider two strategies: \textbf{Abstract-only} classification determines relevance from titles and abstracts; \textbf{Adaptive browsing} permits an ``uncertain'' label for ambiguous cases, triggering full-text examination before final classification. The stage outputs selected papers $\mathcal{S}_t$ and feedback $\mathcal{O}_t$, which summarizes the current search outcomes and suggests refinements for the next iteration.

\paragraph{Memory Mechanism.}
Long-horizon search requires coherent state across iterations. We design two structures: the \textbf{subquery tree} $\mathcal{M}_t$ organizes subqueries hierarchically, recording derivation paths and retrieved papers for each node; the \textbf{experience buffer} $\mathcal{B}_t$ compresses search history into a fixed-length summary, preventing context overflow while preserving key insights.

\subsection{Benchmark Construction}
\label{sec:dataset}

We construct \sysname from two established academic retrieval datasets: PaSa~\cite{he2025pasa} and LitSearch~\cite{ajith2024litsearch}. Queries originate from three splits: PaSa-AutoScholar (generated from citation contexts), PaSa-RealScholar (human-curated research questions), and LitSearch (real-world literature search scenarios). Each query is paired with expert-annotated ground-truth papers. For LitSearch queries lacking explicit temporal bounds, we assign date constraints matching the latest publication date among ground-truth papers. Papers from both datasets are aggregated, deduplicated by arXiv identifier, and enriched with metadata via the arXiv API, yielding a corpus $\mathcal{D}$ of 570K papers spanning computer science, physics, and mathematics.

\paragraph{Evaluation Subsets.} The benchmark is partitioned into two subsets. \textbf{Test-Fast} contains 200 queries sampled for balanced coverage across sources, enabling rapid iteration during development. \textbf{Test-Hard} contains 100 queries on which all evaluated models perform poorly; these queries tend to have larger ground-truth sets (average 2.6 papers) and require finding papers across multiple research areas. Table~\ref{tab:dataset_stats} summarizes dataset statistics; Appendix~\ref{sec:dataset_details} provides detailed construction procedures.

\begin{table}[t]
\caption{Dataset statistics. \#GT denotes average ground-truth papers per query; Len. denotes query length in characters.}
\label{tab:dataset_stats}
\centering
\small
\begin{tabular}{@{}lccc@{}}
\toprule
\textbf{Subset} & \textbf{\#Query} & \textbf{Avg. \#GT} & \textbf{Avg. Len.} \\
\midrule
Test-Fast & 200 & 1.9 & 113.0 \\
Test-Hard & 100 & 2.6 & 101.8 \\
\midrule
ALL & 2,536 & 2.3 & 110.4 \\
\bottomrule
\end{tabular}
\vskip -0.1in
\end{table}

\section{Experiments}
\label{sec:experiments}

\newtcolorbox{findingbox}[1][]{
  colback=blue!5,
  colframe=blue!40!black,
  coltitle=white,
  fonttitle=\bfseries\small,
  title={#1},
  boxrule=0.4pt,
  arc=1.5pt,
  left=5pt,
  right=5pt,
  top=3pt,
  bottom=3pt
}

\subsection{Experimental Setup}
\label{sec:setup}

\paragraph{Backbone Models.}
We evaluate both open-source and proprietary LLMs. Open-source models include Qwen3~\citep{yang2025qwen3} (8B, 30B) and GLM-4.7~\citep{zeng2025glm}. Proprietary models include DeepSeek-V3.2~\citep{liu2025deepseek}, GPT-5.2~\citep{gpt52}, and Gemini3-Pro~\citep{gemini3}. For models that support optional extended thinking (Qwen3 and DeepSeek-V3.2), we evaluate both standard and thinking-enabled configurations; the latter are denoted with $\dagger$.

\paragraph{Baselines.}
We compare against a \textbf{Direct Query} baseline that bypasses iterative query planning entirely. Under this baseline, the original research query $q$ is submitted once to the retrieval backend without decomposition; retrieved candidates then undergo the same relevance assessment procedure as the full workflow.

\paragraph{Retrieval Backend.}
We implement two retrieval methods over paper titles and abstracts. (1) \textbf{Sparse retrieval} employs BM25, which scores documents by term frequency-inverse document frequency weighting and excels at exact lexical matching. (2) \textbf{Dense retrieval} encodes documents using Qwen3-Embedding-0.6B and performs approximate nearest neighbor search in a Qdrant vector database, capturing semantic similarity beyond surface-level term overlap. Unless otherwise noted, sparse retrieval serves as the default backend.

\paragraph{Evaluation Metrics.}
Let $\mathcal{G}$ denote the ground-truth set, $\mathcal{R}$ the retrieved candidates, and $\mathcal{S}$ the final selected papers. We report metrics at two stages: \textit{retrieval} (prefixed ``Ret.'') measures candidate quality before filtering; \textit{selection} evaluates end-to-end performance. Recall and precision are defined as:
\begin{equation}
\text{R} = \frac{|\mathcal{S} \cap \mathcal{G}|}{|\mathcal{G}|}, \quad
\text{P} = \frac{|\mathcal{S} \cap \mathcal{G}|}{|\mathcal{S}|}
\end{equation}

\begin{equation}
\text{Ret.R} = \frac{|\mathcal{R} \cap \mathcal{G}|}{|\mathcal{G}|}, \quad
\text{Ret.P} = \frac{|\mathcal{R} \cap \mathcal{G}|}{|\mathcal{R}|}
\end{equation}

F1 is the harmonic mean of R and P. We additionally introduce two diagnostic metrics:
\begin{itemize}[leftmargin=*,itemsep=2pt,topsep=3pt]
\item \textbf{Avg.Distance} quantifies query planning quality. For each ground-truth paper $g \in \mathcal{G}$, let $r_g$ denote its best rank across all subqueries. With cutoff $c{=}100$:
\begin{equation}
\text{Avg.Dist} = \frac{1}{|\mathcal{G}|} \sum_{g \in \mathcal{G}} \max\left(1 - \frac{r_g}{c}, 0\right)
\end{equation}
Higher values indicate relevant papers are surfaced earlier in retrieval results.
\item \textbf{GT Discard Rate} measures assessment errors---the fraction of discarded candidates that were actually relevant:
\begin{equation}
\text{GT Disc.} = \frac{|(\mathcal{R} \cap \mathcal{G}) \setminus \mathcal{S}|}{|\mathcal{R} \setminus \mathcal{S}|}
\end{equation}
\end{itemize}

\paragraph{Implementation.}
We use Test-Fast with $T{=}5$ iterations and greedy decoding (Temperature=0, Top\_p=1) for reproducibility. Main results (Table~\ref{tab:main_results}) report metrics at the final iteration under sparse retrieval and Abstract-only assessment mode.

\subsection{Main Results}
\label{sec:main_results}

\begin{table*}[t]
\caption{Performance on Test-Fast and Test-Hard benchmarks (sparse retrieval, Abstract-only, iteration 5). R/P/F1 denote selection-stage recall, precision, and F1; Ret.R/Ret.P/Ret.F1 denote retrieval-stage metrics. Best results are \textbf{bolded}; second-best are \underline{underlined}. $\dagger$ indicates extended thinking mode.}
\label{tab:main_results}
\centering
\small
\setlength{\tabcolsep}{3.2pt}
\begin{tabular}{@{}l ccc ccc | ccc ccc@{}}
\toprule
& \multicolumn{6}{c|}{\textbf{Test-Fast}} & \multicolumn{6}{c}{\textbf{Test-Hard}} \\
\cmidrule(lr){2-7} \cmidrule(lr){8-13}
\textbf{Model} & \textbf{R} & \textbf{P} & \textbf{F1} & \textbf{Ret.R} & \textbf{Ret.P} & \textbf{Ret.F1} & \textbf{R} & \textbf{P} & \textbf{F1} & \textbf{Ret.R} & \textbf{Ret.P} & \textbf{Ret.F1} \\
\midrule
\multicolumn{13}{@{}l}{\textit{Direct Query Baseline}} \\
\quad Qwen3-8B & 0.185 & 0.042 & 0.069 & 0.210 & 0.036 & 0.061 & 0.048 & 0.004 & 0.007 & 0.055 & 0.003 & 0.006 \\
\quad Qwen3-30B & 0.312 & 0.058 & 0.098 & 0.350 & 0.052 & 0.091 & 0.089 & 0.006 & 0.011 & 0.095 & 0.005 & 0.010 \\
\midrule
\multicolumn{13}{@{}l}{\textit{Open-Source Models}} \\
\quad Qwen3-8B & 0.483 & 0.152 & 0.231 & 0.550 & 0.022 & 0.042 & 0.123 & 0.016 & 0.028 & 0.165 & 0.005 & 0.010 \\
\quad Qwen3-8B$^\dagger$ & 0.458 & 0.216 & 0.293 & 0.580 & \underline{0.023} & \underline{0.045} & 0.120 & 0.013 & 0.023 & 0.213 & 0.007 & 0.014 \\
\quad Qwen3-30B & 0.673 & 0.181 & 0.285 & 0.720 & 0.011 & 0.021 & 0.226 & 0.016 & 0.030 & 0.316 & 0.006 & 0.011 \\
\quad Qwen3-30B$^\dagger$ & 0.482 & \underline{0.290} & 0.362 & 0.590 & \textbf{0.025} & \textbf{0.048} & 0.172 & \textbf{0.058} & \textbf{0.087} & 0.264 & \underline{0.010} & \underline{0.019} \\
\quad GLM-4.7 & 0.754 & 0.111 & 0.194 & 0.814 & 0.010 & 0.020 & 0.216 & 0.017 & 0.031 & 0.398 & 0.006 & 0.012 \\
\midrule
\multicolumn{13}{@{}l}{\textit{Proprietary Models}} \\
\quad DeepSeek-V3.2 & \underline{0.855} & 0.135 & 0.233 & 0.862 & 0.007 & 0.014 & \underline{0.372} & 0.017 & 0.032 & 0.430 & 0.006 & 0.012 \\
\quad DeepSeek-V3.2$^\dagger$ & 0.812 & 0.287 & \underline{0.423} & 0.872 & 0.009 & 0.018 & 0.355 & 0.041 & 0.074 & 0.448 & 0.008 & 0.016 \\
\quad GPT-5.2 & 0.837 & \textbf{0.305} & \textbf{0.447} & \underline{0.883} & 0.009 & 0.018 & \textbf{0.397} & \underline{0.045} & \underline{0.081} & \textbf{0.505} & 0.008 & 0.016 \\
\quad Gemini3-Pro & \textbf{0.950} & 0.199 & 0.329 & \textbf{0.958} & 0.011 & 0.022 & 0.365 & 0.028 & 0.052 & \underline{0.504} & \textbf{0.010} & \textbf{0.020} \\
\bottomrule
\end{tabular}
\end{table*}

Table~\ref{tab:main_results} presents performance on both Test-Fast and Test-Hard benchmarks.

\paragraph{Iterative Planning Effectiveness.}
Comparison with the Direct Query baseline quantifies the contribution of iterative decomposition. Without query planning, Qwen3-30B achieves only 0.098 F1; with the full workflow, F1 reaches 0.285---a 2.9$\times$ improvement. For Qwen3-8B, the gain is 3.3$\times$ (0.069 to 0.231). Single-query retrieval fails to cover the semantic breadth of complex research questions that span multiple methodological and application domains.

\paragraph{Overall Performance.}
GPT-5.2 achieves the highest F1 (0.447) by balancing recall (0.837) and precision (0.305). Gemini3-Pro attains the highest recall (0.950) but at lower precision (0.199), yielding F1 of 0.329. A substantial performance gap separates proprietary and open-source models: the best open-source result (Qwen3-30B$^\dagger$, F1=0.362) trails GPT-5.2 by 19\% in relative terms. We analyze the sources of this gap in Section~\ref{sec:component_analysis}.

\paragraph{Retrieval vs.\ Selection Performance.}
Comparing Ret.R and R reveals where ground-truth coverage is lost during assessment. Gemini3-Pro retrieves 95.8\% of ground-truth papers and retains 95.0\% after assessment---a gap of only 0.8 percentage points. In contrast, Qwen3-8B$^\dagger$ exhibits a 12.2 point drop (Ret.R=0.580 to R=0.458) due to aggressive filtering. This gap is inversely correlated with model capability, indicating that relevance assessment represents a high-leverage optimization target for smaller models.

\paragraph{Extended Thinking Trade-offs.}
Extended thinking induces a precision-recall trade-off. Qwen3-8B$^\dagger$ achieves 42\% higher precision than its base variant (0.216 vs.\ 0.152) while sacrificing 5\% recall (0.458 vs.\ 0.483). This trade-off arises from more aggressive filtering during relevance assessment, which we analyze in Section~\ref{sec:assessment_analysis}. The net F1 improvement scales with model capability: DeepSeek-V3.2$^\dagger$ gains +82\% F1 over DeepSeek-V3.2 (0.423 vs.\ 0.233), whereas Qwen3-8B$^\dagger$ gains +27\% (0.293 vs.\ 0.231).

\paragraph{Model Scaling.}
Within the Qwen3 family, scaling from 8B to 30B yields substantial recall improvement (+39\%, from 0.483 to 0.673) but only marginal precision gain (0.181 vs.\ 0.152). Larger models generate more diverse subqueries that cover broader semantic regions, yet do not necessarily impose stricter relevance criteria during assessment.

\paragraph{Generalization to Test-Hard.}
All models degrade substantially on Test-Hard, which comprises cross-domain queries with larger ground-truth sets. The best Test-Fast F1 (0.447) drops to 0.087---an 80\% decline. Notably, Qwen3-30B$^\dagger$ achieves the highest Test-Hard F1 (0.087), marginally outperforming GPT-5.2 (0.081). However, this advantage derives from higher precision (0.058 vs.\ 0.045) rather than recall; GPT-5.2 maintains the highest Test-Hard recall (0.397 vs.\ 0.172). The relative F1 gain of thinking-enabled models on cross-disciplinary queries reflects their conservative selection strategy rather than improved discovery coverage.

\begin{findingbox}[Key Findings]
(1) Iterative planning improves F1 by 2.9--3.3$\times$ over Direct Query. (2) Gemini3-Pro achieves the highest recall (0.950); GPT-5.2 achieves the best F1 (0.447) by balancing recall and precision. (3) Extended thinking trades recall for precision via more aggressive filtering; F1 gains scale with model capability.
\end{findingbox}

\subsection{Iteration Dynamics}
\label{sec:iteration_analysis}

\begin{table}[t]
\caption{Per-iteration recall and precision on Test-Fast (sparse retrieval, Abstract-only).}
\label{tab:iteration_metrics}
\centering
\small
\setlength{\tabcolsep}{3.2pt}
\begin{tabular}{@{}l ccccc@{}}
\toprule
\textbf{Model} & \textbf{It.1} & \textbf{It.2} & \textbf{It.3} & \textbf{It.4} & \textbf{It.5} \\
\midrule
\multicolumn{6}{@{}l}{\textit{Recall}} \\
\quad Qwen3-8B & 0.332 & 0.403 & 0.440 & 0.453 & 0.483 \\
\quad Qwen3-8B$^\dagger$ & 0.298 & 0.400 & 0.435 & 0.450 & 0.458 \\
\quad Qwen3-30B & 0.526 & 0.626 & 0.659 & 0.669 & 0.673 \\
\quad Qwen3-30B$^\dagger$ & 0.375 & 0.410 & 0.451 & 0.460 & 0.482 \\
\quad GLM-4.7 & 0.515 & 0.602 & 0.659 & 0.714 & 0.754 \\
\quad DeepSeek-V3.2 & 0.540 & 0.655 & 0.745 & 0.810 & 0.855 \\
\quad DeepSeek-V3.2$^\dagger$ & 0.498 & 0.615 & 0.710 & 0.775 & 0.812 \\
\quad GPT-5.2 & 0.618 & 0.725 & 0.772 & 0.805 & 0.837 \\
\quad Gemini3-Pro & 0.694 & 0.812 & 0.882 & 0.925 & 0.950 \\
\midrule
\multicolumn{6}{@{}l}{\textit{Precision}} \\
\quad Qwen3-8B & 0.177 & 0.149 & 0.144 & 0.148 & 0.152 \\
\quad Qwen3-8B$^\dagger$ & 0.218 & 0.245 & 0.241 & 0.227 & 0.216 \\
\quad Qwen3-30B & 0.227 & 0.195 & 0.185 & 0.185 & 0.181 \\
\quad Qwen3-30B$^\dagger$ & 0.292 & 0.291 & 0.295 & 0.290 & 0.290 \\
\quad GLM-4.7 & 0.182 & 0.158 & 0.136 & 0.115 & 0.111 \\
\quad DeepSeek-V3.2 & 0.220 & 0.172 & 0.148 & 0.141 & 0.135 \\
\quad DeepSeek-V3.2$^\dagger$ & 0.395 & 0.341 & 0.311 & 0.297 & 0.287 \\
\quad GPT-5.2 & 0.345 & 0.328 & 0.315 & 0.310 & 0.305 \\
\quad Gemini3-Pro & 0.229 & 0.199 & 0.183 & 0.199 & 0.199 \\
\bottomrule
\end{tabular}
\end{table}

\begin{figure}[t]
\centering
\includegraphics[width=1\linewidth]{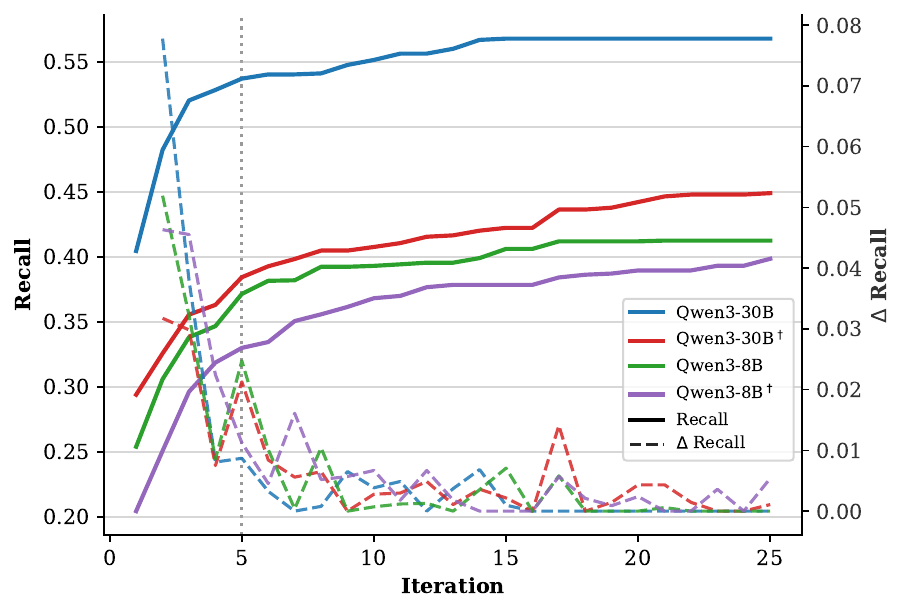}
\caption{Recall trajectories over 25 iterations (Qwen3 models). Dashed line: default $T{=}5$. Solid lines show cumulative recall (left axis); dashed lines show per-iteration $\Delta$Recall (right axis).}
\label{fig:extended_iter}
\vspace{-3mm}
\end{figure}

Table~\ref{tab:iteration_metrics} presents per-iteration metrics, while Figure~\ref{fig:extended_iter} extends the analysis to 25 iterations for detailed convergence characterization.

\paragraph{Convergence Patterns.}
Two distinct regimes emerge across model capabilities. High-capability models front-load discovery: Gemini3-Pro achieves 73\% of its final recall (0.694/0.950) in iteration 1, with diminishing marginal gains of +12\%, +7\%, +4\%, and +3\% in subsequent iterations---reflecting effective initial query decomposition. Lower-capability models exhibit more gradual convergence: Qwen3-8B gains +21\%, +9\%, +3\%, and +7\% across iterations 2--5, deriving greater benefit from feedback-driven query refinement.

\paragraph{Precision Dynamics.}
Precision decays monotonically as iterations accumulate candidates. DeepSeek-V3.2$^\dagger$ decreases from 0.395 to 0.287 ($-$27\%); GLM-4.7 from 0.182 to 0.111 ($-$39\%). A notable exception is Qwen3-30B$^\dagger$, which maintains stable precision (0.292 to 0.290) throughout the iteration sequence. This stability suggests that thinking-enabled assessment enforces consistent relevance thresholds irrespective of candidate volume, whereas standard models progressively lower their acceptance criteria.

\paragraph{Saturation Beyond $T{=}5$.}
Figure~\ref{fig:extended_iter} demonstrates that all Qwen3 variants plateau within 10 iterations. Beyond $T{=}5$, Qwen3-30B gains only 2.3 percentage points (0.673 to 0.702); Qwen3-8B gains merely 0.4 points (0.483 to 0.487). Thinking-enabled variants exhibit analogous saturation patterns. This ceiling reflects exhaustion of productive query reformulations---later iterations revisit previously explored semantic regions rather than discovering novel relevant work. These observations motivate our default setting of $T{=}5$ as an effective trade-off between coverage and computational cost.

\begin{findingbox}[Key Findings]
(1) High-capability models achieve 70--90\% of final recall within two iterations. (2) Precision decays with iterations for most models; thinking-enabled variants maintain more stable precision. (3) Recall saturates by $T{=}10$; gains beyond $T{=}5$ are minimal.
\end{findingbox}

\subsection{Component Analysis}
\label{sec:component_analysis}

\subsubsection{Query Planning Quality}
\label{sec:planning_analysis}

\begin{table}[t]
\caption{Avg.Distance by iteration on Test-Fast (sparse retrieval). Higher values indicate ground-truth papers rank earlier in retrieval results.}
\label{tab:planning_quality}
\centering
\small
\setlength{\tabcolsep}{3.5pt}
\begin{tabular}{@{}l ccccc c@{}}
\toprule
\textbf{Model} & \textbf{It.1} & \textbf{It.2} & \textbf{It.3} & \textbf{It.4} & \textbf{It.5} & \textbf{Avg.} \\
\midrule
Qwen3-8B & 0.588 & 0.600 & 0.604 & 0.587 & 0.583 & 0.592 \\
Qwen3-8B$^\dagger$ & 0.582 & 0.626 & 0.626 & 0.623 & 0.616 & 0.615 \\
Qwen3-30B & 0.695 & 0.735 & 0.731 & 0.702 & 0.708 & 0.714 \\
Qwen3-30B$^\dagger$ & 0.668 & 0.685 & 0.678 & 0.672 & 0.667 & 0.674 \\
GLM-4.7 & 0.730 & 0.771 & 0.775 & 0.798 & 0.813 & 0.777 \\
\midrule
DeepSeek-V3.2 & 0.739 & 0.773 & 0.809 & 0.835 & 0.845 & 0.800 \\
DeepSeek-V3.2$^\dagger$ & 0.737 & 0.800 & 0.831 & 0.820 & 0.816 & 0.801 \\
GPT-5.2 & 0.771 & 0.803 & 0.820 & 0.826 & 0.836 & 0.811 \\
Gemini3-Pro & 0.784 & 0.828 & 0.855 & 0.876 & \textbf{0.881} & \textbf{0.845} \\
\bottomrule
\end{tabular}
\end{table}

Table~\ref{tab:planning_quality} presents Avg.Distance trajectories across iterations.

\paragraph{Model Hierarchy.}
Gemini3-Pro attains the highest Avg.Distance (0.845), with subqueries ranking ground-truth papers within the top 15--20 positions on average. A substantial gap separates proprietary models (0.800--0.845) from open-source alternatives (0.592--0.777). This 43\% disparity between Gemini3-Pro and Qwen3-8B accounts for much of the recall difference observed in Table~\ref{tab:main_results}: well-formulated queries surface relevant papers earlier in the ranking, increasing discovery probability within the fixed retrieval budget.

\paragraph{Iteration Dynamics.}
Planning quality improves across iterations through feedback incorporation. Gemini3-Pro increases from 0.784 (iteration 1) to 0.881 (iteration 5), representing a 12\% gain. Proprietary models exhibit steeper improvement trajectories (+10\% on average) compared to open-source counterparts (+3\%), suggesting more effective utilization of feedback signals from prior retrieval outcomes.

\begin{findingbox}[Key Findings]
(1) Proprietary models achieve 43\% higher Avg.Distance than open-source alternatives. (2) Planning quality improves across iterations via feedback incorporation, with proprietary models showing steeper gains (+10\% vs.\ +3\%).
\end{findingbox}

\subsubsection{Relevance Assessment}
\label{sec:assessment_analysis}

\begin{figure}[t]
\centering
\includegraphics[width=0.95\linewidth]{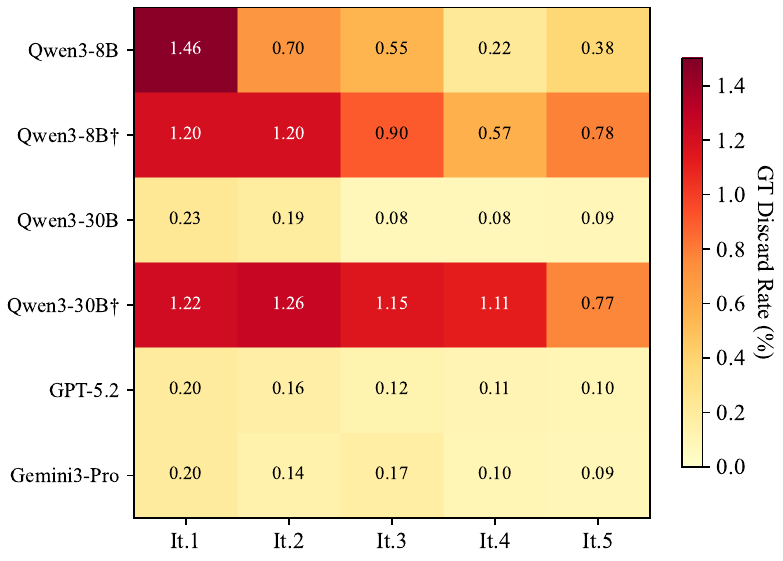}
\caption{Per-iteration GT Discard Rate (\%) on Test-Fast. Darker cells indicate higher discard rates of ground-truth papers during relevance assessment.}
\label{fig:discard_heatmap}
\vspace{-3mm}
\end{figure}

\begin{table}[t]
\caption{GT Discard Rate (\%) and precision by assessment strategy on Test-Fast.}
\label{tab:assessment_discard}
\centering
\small
\begin{tabular}{@{}l cc cc@{}}
\toprule
& \multicolumn{2}{c}{\textbf{Abstract-only}} & \multicolumn{2}{c}{\textbf{Adaptive}} \\
\cmidrule(lr){2-3} \cmidrule(lr){4-5}
\textbf{Model} & \textbf{Disc.\%} & \textbf{Prec.} & \textbf{Disc.\%} & \textbf{Prec.} \\
\midrule
Qwen3-8B & 0.66 & 0.152 & 0.34 & 0.149 \\
Qwen3-8B$^\dagger$ & 0.93 & 0.216 & 0.58 & 0.218 \\
Qwen3-30B & 0.13 & 0.181 & 0.29 & 0.217 \\
Qwen3-30B$^\dagger$ & 1.03 & 0.290 & 0.71 & 0.311 \\
GLM-4.7 & 0.36 & 0.111 & 0.23 & 0.128 \\
\midrule
DeepSeek-V3.2 & 0.12 & 0.135 & 0.10 & 0.167 \\
DeepSeek-V3.2$^\dagger$ & 0.17 & 0.287 & 0.12 & 0.302 \\
GPT-5.2 & 0.14 & 0.305 & 0.10 & 0.320 \\
Gemini3-Pro & 0.14 & 0.199 & 0.11 & 0.232 \\
\bottomrule
\end{tabular}
\end{table}

Figure~\ref{fig:discard_heatmap} and Table~\ref{tab:assessment_discard} characterize relevance assessment behavior.

\paragraph{Error Patterns.}
Proprietary models maintain GT Discard Rates below 0.20\% across all iterations (Figure~\ref{fig:discard_heatmap}), with errors distributed randomly. Thinking-enabled open-source models exhibit elevated, iteration-dependent rates: Qwen3-8B$^\dagger$ starts at 1.46\% in iteration 1 and declines to 0.38\% by iteration 5; Qwen3-30B$^\dagger$ remains consistently elevated (1.22\% to 0.77\%). This trajectory indicates that extended thinking applies aggressive relevance thresholds that gradually calibrate through accumulated assessment experience.

\paragraph{Precision-Error Trade-off.}
Higher GT Discard Rates correlate with elevated precision, quantifying the cost of stringent filtering. Qwen3-30B$^\dagger$ achieves the highest open-source precision (0.290) alongside the highest discard rate (1.03\%), whereas Qwen3-30B maintains minimal discard rates (0.13\%) with lower precision (0.181). This pattern explains the precision-recall trade-off observed in Section~\ref{sec:main_results}: thinking-enabled models improve precision through aggressive filtering at the cost of discarding some relevant papers.

\paragraph{Adaptive Browsing.}
Full-text retrieval for ambiguous candidates reduces GT Discard Rates while preserving precision. Qwen3-8B$^\dagger$ improves from 0.93\% to 0.58\%; Qwen3-30B$^\dagger$ from 1.03\% to 0.71\%. Proprietary models show negligible change, as their abstract-based assessment already achieves high accuracy. The benefit is most pronounced for methodology papers where relevance depends on technical details absent from abstracts.

\begin{findingbox}[Key Findings]
(1) Thinking-enabled models achieve higher precision via aggressive filtering (GT Discard Rates: 0.93--1.03\% vs.\ $<$0.20\% for proprietary models). (2) Adaptive Browsing reduces discard rates by 30--40\% for open-source models.
\end{findingbox}

\subsection{Ablation Studies}
\label{sec:ablation}

\subsubsection{Retrieval Backend}
\label{sec:ablation_retrieval}

\begin{table}[t]
\caption{Sparse versus dense retrieval on Test-Fast (Abstract-only, iteration 5). Dense retrieval uses Qwen3-Embedding-0.6B for document encoding.}
\label{tab:ablation_retrieval}
\centering
\small
\setlength{\tabcolsep}{4pt}
\begin{tabular}{@{}l ccc ccc@{}}
\toprule
& \multicolumn{3}{c}{\textbf{Sparse}} & \multicolumn{3}{c}{\textbf{Dense}} \\
\cmidrule(lr){2-4} \cmidrule(lr){5-7}
\textbf{Model} & \textbf{R} & \textbf{P} & \textbf{F1} & \textbf{R} & \textbf{P} & \textbf{F1} \\
\midrule
Qwen3-8B & 0.483 & 0.152 & 0.231 & 0.608 & 0.183 & 0.281 \\
Qwen3-8B$^\dagger$ & 0.458 & 0.216 & 0.293 & 0.488 & 0.209 & 0.293 \\
Qwen3-30B & 0.673 & 0.181 & 0.285 & 0.718 & 0.208 & 0.322 \\
Qwen3-30B$^\dagger$ & 0.482 & 0.290 & 0.362 & 0.532 & 0.287 & 0.373 \\
\bottomrule
\end{tabular}
\end{table}

Table~\ref{tab:ablation_retrieval} compares both sparse and dense retrieval backends.

\paragraph{Sparse vs.\ Dense Retrieval.}
Dense retrieval yields model-dependent recall improvements. Standard models exhibit substantial gains: Qwen3-8B improves by 12.5 percentage points (0.483 to 0.608), and Qwen3-30B by 4.5 points (0.673 to 0.718). Thinking-enabled models show smaller improvements (3.0--5.0 points), likely because their stricter assessment thresholds limit the net recall gain even when additional relevant candidates are surfaced.

\begin{findingbox}[Key Findings]
Dense retrieval improves recall by 7--26\% for standard models, with smaller gains (7--10\%) for thinking-enabled variants due to stricter assessment filtering.
\end{findingbox}

\subsubsection{Memory Mechanism}
\label{sec:ablation_memory}

\begin{table}[t]
\caption{Effect of the memory mechanism on Test-Fast performance. ``w/o Memory'' removes the experience buffer $\mathcal{B}_t$, retaining only raw conversation history as context input to the Query Planner.}
\label{tab:ablation_memory}
\centering
\small
\begin{tabular}{@{}l ccc ccc@{}}
\toprule
& \multicolumn{3}{c}{\textbf{With Memory}} & \multicolumn{3}{c}{\textbf{w/o Memory}} \\
\cmidrule(lr){2-4} \cmidrule(lr){5-7}
\textbf{Model} & \textbf{R} & \textbf{P} & \textbf{F1} & \textbf{R} & \textbf{P} & \textbf{F1} \\
\midrule
Qwen3-8B & 0.483 & 0.152 & 0.231 & 0.485 & 0.140 & 0.217 \\
Qwen3-8B$^\dagger$ & 0.458 & 0.216 & 0.293 & 0.411 & 0.168 & 0.239 \\
Qwen3-30B & 0.673 & 0.181 & 0.285 & 0.651 & 0.169 & 0.268 \\
Qwen3-30B$^\dagger$ & 0.482 & 0.290 & 0.362 & 0.402 & 0.218 & 0.282 \\
\bottomrule
\end{tabular}
\end{table}

Table~\ref{tab:ablation_memory} quantifies the contribution of the memory mechanism.

\paragraph{Performance Impact.}
The experience buffer $\mathcal{B}_t$ aggregates the reasoning traces and outputs from all prior query planning iterations into a compressed summary, tracking discovered papers and explored query regions to guide subsequent iterations toward unexplored areas. Removing this component degrades performance uniformly across models. Thinking-enabled models exhibit the largest drops: Qwen3-8B$^\dagger$ decreases from 0.293 to 0.239 F1 ($-$18.4\%), and Qwen3-30B$^\dagger$ from 0.362 to 0.282 ($-$22.1\%). Standard models show more modest degradation (6.0--6.1\%).

\paragraph{Compression vs.\ Raw History.}
Without the experience buffer, query planning receives raw conversation history instead of a compressed summary. The lengthier, unstructured context impedes effective extraction of exploration state, leading to regeneration of semantically similar subqueries and wasted retrieval budget. Thinking-enabled models are disproportionately affected: their verbose reasoning traces produce longer histories that exacerbate context utilization difficulties, whereas the compressed buffer distills prior exploration into an efficiently consumable format.

\begin{findingbox}[Key Findings]
The memory mechanism prevents query redundancy across iterations. Removing $\mathcal{B}_t$ degrades F1 by 6--22\%, with thinking-enabled models most affected.
\end{findingbox}

\section{Conclusion}

We introduced \sysname to address the reproducibility challenge in deep research evaluation. By grounding agentic workflows in a static, deterministic environment containing 570K papers, we isolated algorithmic reasoning from live API variance, enabling precise dissection of model behaviors. Our modular analysis reveals distinct operating regimes: while proprietary models maximize discovery through superior query formulation, ``extended thinking'' paradigms in open models function primarily as high-precision filters, often at the cost of recall. We further identified query planning quality and relevance assessment calibration as the dual bottlenecks limiting open-source performance on complex tasks.

Our findings position \sysname as a critical testbed for next-generation research agents. The deterministic environment specifically enables the training of reinforcement learning policies for query planning without the variance of live web search, paving the way for agents that can learn optimal exploration-exploitation trade-offs in information seeking. We open-source our environment and benchmarks to facilitate this community effort.

\section*{Impact Statement}

This paper presents work whose goal is to advance the field of Machine
Learning. There are many potential societal consequences of our work, none
which we feel must be specifically highlighted here.

\bibliography{example_paper}
\bibliographystyle{icml2026}

\newpage
\appendix
\onecolumn

\section{Additional Experimental Results}
\label{sec:additional_results}

\subsection{Additional Visualizations}
\label{sec:additional_viz}

\begin{figure}[htbp]
\centering
\includegraphics[width=1.0\linewidth]{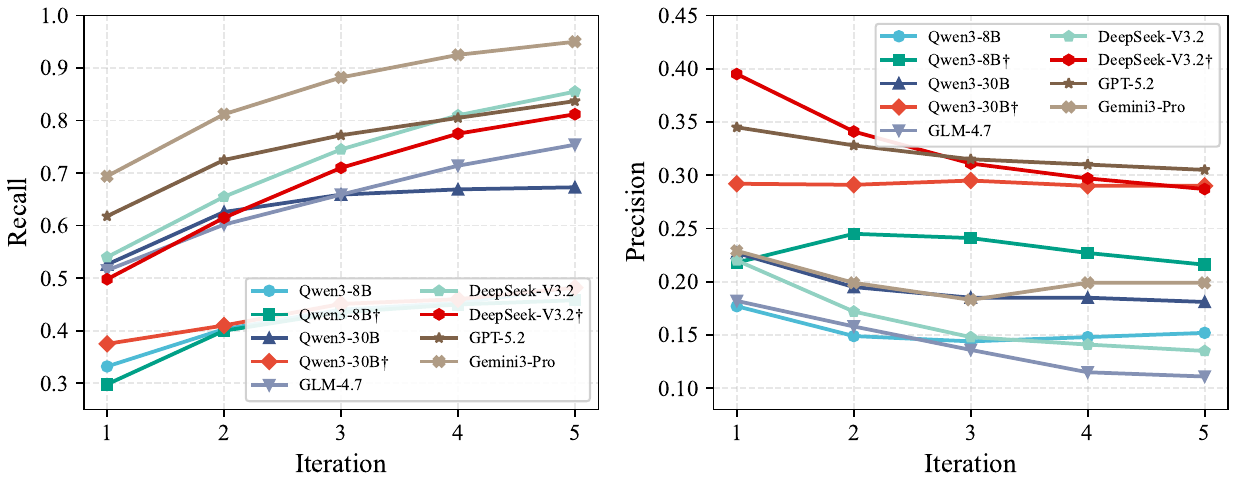}
\caption{Recall and Precision trajectories across 5 iterations on Test-Fast (sparse retrieval, Abstract-only).}
\label{fig:iteration_curves_app}
\end{figure}

\begin{figure}[htbp]
\centering
\includegraphics[width=0.6\linewidth]{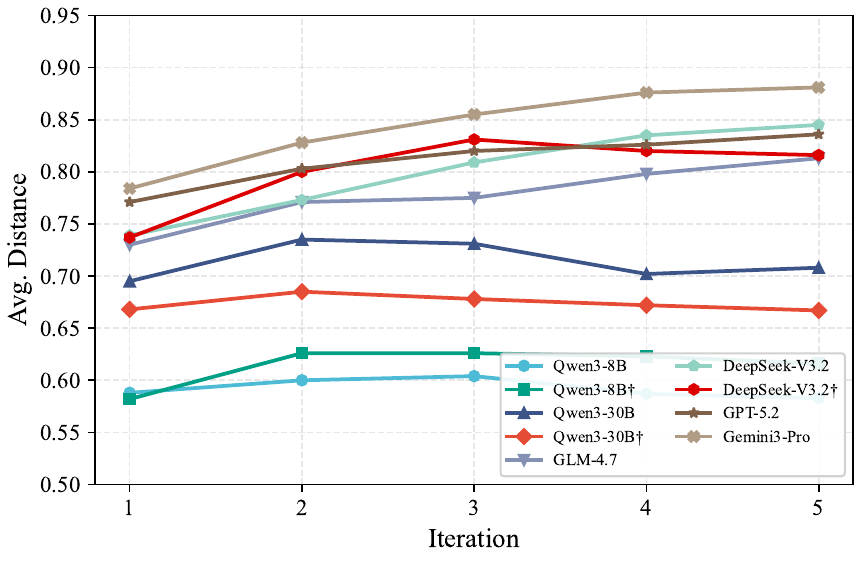}
\caption{Avg.Distance trajectories across iterations. Higher values indicate queries that rank ground-truth papers earlier in retrieval results.}
\label{fig:planning_distance_app}
\end{figure}

\begin{figure}[htbp]
\centering
\includegraphics[width=0.5\linewidth]{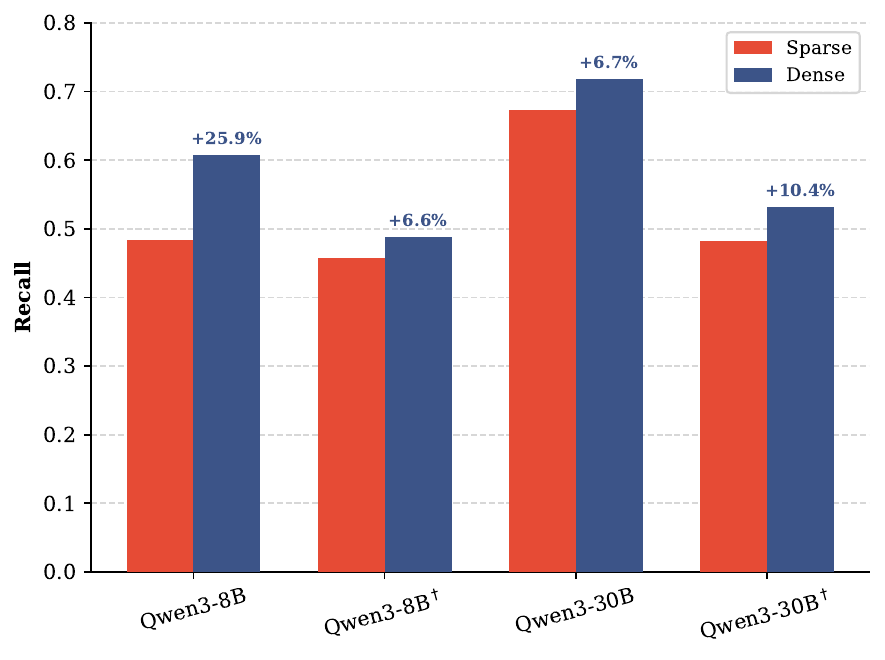}
\caption{Sparse versus dense retrieval recall comparison.}
\label{fig:retrieval_comparison_app}
\end{figure}

\begin{figure}[htbp]
\centering
\includegraphics[width=0.5\linewidth]{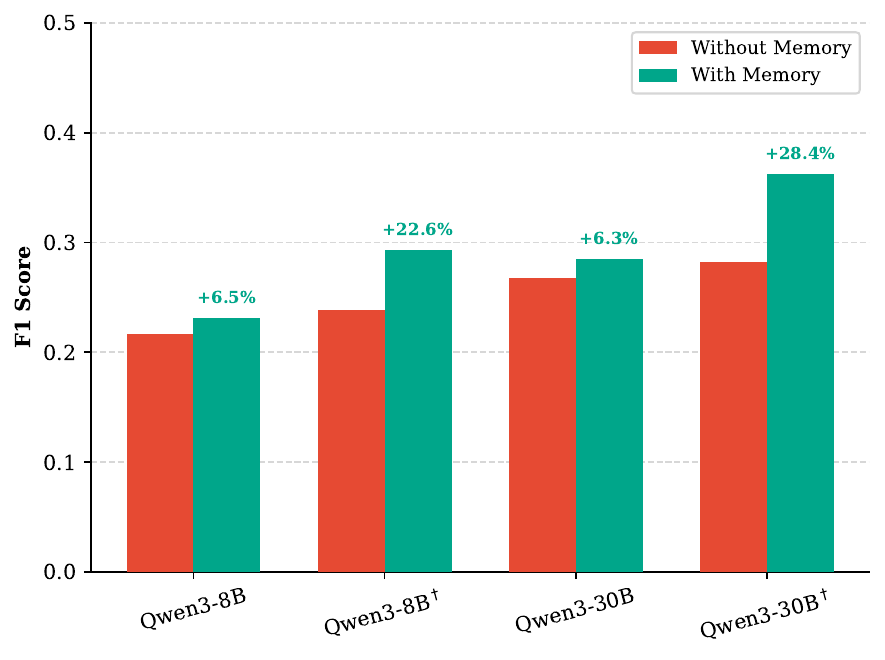}
\caption{Impact of the memory mechanism. Removing the experience buffer $\mathcal{B}_t$ degrades F1 across all models, with percentage drops annotated.}
\label{fig:memory_ablation_app}
\end{figure}

Figure~\ref{fig:iteration_curves_app} presents detailed recall and precision trajectories across iterations. Figure~\ref{fig:planning_distance_app} shows query planning quality trajectories across iterations. Figures~\ref{fig:retrieval_comparison_app} and~\ref{fig:memory_ablation_app} present ablation study results.

\subsection{Detailed Test-Hard Results}
\label{sec:test_hard_details}

Table~\ref{tab:test_hard_details} presents comprehensive performance metrics on the Test-Hard benchmark. Test-Hard queries span multiple research domains and exhibit higher complexity than Test-Fast, resulting in substantially lower performance across all models.

\begin{table*}[htbp]
\caption{Detailed performance on Test-Hard benchmark (sparse retrieval, Abstract-only, iteration~5). R/P/F1 denote selection-stage metrics; Ret.R/Ret.P/Ret.F1 denote retrieval-stage metrics. Avg.Dist measures query planning effectiveness.}
\label{tab:test_hard_details}
\centering
\small
\setlength{\tabcolsep}{4pt}
\begin{tabular}{@{}l cccc ccc@{}}
\toprule
\textbf{Model} & \textbf{Avg.Dist} & \textbf{R} & \textbf{P} & \textbf{F1} & \textbf{Ret.R} & \textbf{Ret.P} & \textbf{Ret.F1} \\
\midrule
\multicolumn{8}{@{}l}{\textit{Open-Source Models}} \\
\quad Qwen3-8B & 0.270 & 0.123 & 0.016 & 0.028 & 0.165 & 0.005 & 0.010 \\
\quad Qwen3-8B$^\dagger$ & 0.258 & 0.120 & 0.013 & 0.023 & 0.213 & 0.007 & 0.014 \\
\quad Qwen3-30B & 0.311 & 0.226 & 0.016 & 0.030 & 0.316 & 0.006 & 0.011 \\
\quad Qwen3-30B$^\dagger$ & 0.264 & 0.172 & 0.058 & 0.087 & 0.264 & 0.010 & 0.019 \\
\quad GLM-4.7 & 0.341 & 0.216 & 0.017 & 0.031 & 0.398 & 0.006 & 0.012 \\
\midrule
\multicolumn{8}{@{}l}{\textit{Proprietary Models}} \\
\quad DeepSeek-V3.2 & 0.387 & 0.372 & 0.017 & 0.032 & 0.430 & 0.007 & 0.014 \\
\quad DeepSeek-V3.2$^\dagger$ & 0.398 & 0.355 & 0.041 & 0.074 & 0.448 & 0.008 & 0.015 \\
\quad GPT-5.2 & 0.383 & 0.397 & 0.045 & 0.081 & 0.505 & 0.008 & 0.016 \\
\quad Gemini3-Pro & 0.404 & 0.365 & 0.028 & 0.052 & 0.504 & 0.011 & 0.020 \\
\bottomrule
\end{tabular}
\end{table*}

\subsection{Cumulative Metrics on Test-Hard}
\label{sec:test_hard_iterations}

Table~\ref{tab:test_hard_iterations} shows per-iteration recall and precision on Test-Hard. Compared to Test-Fast, all models exhibit slower convergence and substantially lower final performance.

\begin{table*}[htbp]
\caption{Cumulative recall and precision at each iteration on Test-Hard (sparse retrieval, Abstract-only).}
\label{tab:test_hard_iterations}
\centering
\small
\setlength{\tabcolsep}{5pt}
\begin{tabular}{@{}l ccccc | ccccc@{}}
\toprule
& \multicolumn{5}{c|}{\textbf{Recall}} & \multicolumn{5}{c}{\textbf{Precision}} \\
\cmidrule(lr){2-6} \cmidrule(lr){7-11}
\textbf{Model} & \textbf{It.1} & \textbf{It.2} & \textbf{It.3} & \textbf{It.4} & \textbf{It.5} & \textbf{It.1} & \textbf{It.2} & \textbf{It.3} & \textbf{It.4} & \textbf{It.5} \\
\midrule
Qwen3-8B & 0.080 & 0.091 & 0.111 & 0.111 & 0.123 & 0.020 & 0.013 & 0.019 & 0.017 & 0.016 \\
Qwen3-8B$^\dagger$ & 0.035 & 0.073 & 0.077 & 0.104 & 0.120 & 0.015 & 0.016 & 0.011 & 0.013 & 0.013 \\
Qwen3-30B & 0.126 & 0.155 & 0.204 & 0.206 & 0.226 & 0.023 & 0.020 & 0.019 & 0.016 & 0.016 \\
Qwen3-30B$^\dagger$ & 0.118 & 0.143 & 0.149 & 0.154 & 0.172 & 0.054 & 0.058 & 0.047 & 0.044 & 0.058 \\
GLM-4.7 & 0.086 & 0.113 & 0.169 & 0.188 & 0.216 & 0.023 & 0.017 & 0.017 & 0.017 & 0.017 \\
\midrule
DeepSeek-V3.2 & 0.235 & 0.290 & 0.330 & 0.355 & 0.372 & 0.027 & 0.022 & 0.019 & 0.018 & 0.017 \\
DeepSeek-V3.2$^\dagger$ & 0.220 & 0.275 & 0.315 & 0.340 & 0.355 & 0.056 & 0.049 & 0.044 & 0.042 & 0.041 \\
GPT-5.2 & 0.285 & 0.338 & 0.365 & 0.382 & 0.397 & 0.052 & 0.048 & 0.047 & 0.046 & 0.045 \\
Gemini3-Pro & 0.265 & 0.310 & 0.337 & 0.352 & 0.365 & 0.033 & 0.028 & 0.026 & 0.028 & 0.028 \\
\bottomrule
\end{tabular}
\end{table*}

\subsection{Impact of Adaptive Browsing on Test-Hard}
\label{sec:browser_hard}

Table~\ref{tab:browser_hard} compares Abstract-only and Adaptive Browsing strategies on Test-Hard. While Adaptive Browsing improves recall for most models, the precision gains are less consistent than on Test-Fast.

\begin{table}[htbp]
\caption{Abstract-only versus Adaptive Browsing on Test-Hard (sparse retrieval, iteration~5).}
\label{tab:browser_hard}
\centering
\small
\setlength{\tabcolsep}{3pt}
\begin{tabular}{@{}l ccc ccc@{}}
\toprule
& \multicolumn{3}{c}{\textbf{Abstract-only}} & \multicolumn{3}{c}{\textbf{Adaptive}} \\
\cmidrule(lr){2-4} \cmidrule(lr){5-7}
\textbf{Model} & \textbf{R} & \textbf{P} & \textbf{F1} & \textbf{R} & \textbf{P} & \textbf{F1} \\
\midrule
Qwen3-8B & 0.123 & 0.016 & 0.028 & 0.188 & 0.026 & 0.046 \\
Qwen3-8B$^\dagger$ & 0.120 & 0.013 & 0.023 & 0.179 & 0.019 & 0.035 \\
Qwen3-30B & 0.226 & 0.016 & 0.030 & 0.256 & 0.035 & 0.061 \\
Qwen3-30B$^\dagger$ & 0.172 & 0.058 & 0.087 & 0.160 & 0.042 & 0.066 \\
GLM-4.7 & 0.216 & 0.017 & 0.031 & 0.317 & 0.023 & 0.042 \\
\midrule
DeepSeek-V3.2 & 0.372 & 0.017 & 0.032 & 0.385 & 0.021 & 0.040 \\
DeepSeek-V3.2$^\dagger$ & 0.355 & 0.041 & 0.074 & 0.368 & 0.043 & 0.078 \\
GPT-5.2 & 0.397 & 0.045 & 0.081 & 0.422 & 0.050 & 0.090 \\
Gemini3-Pro & 0.365 & 0.028 & 0.052 & 0.372 & 0.033 & 0.062 \\
\bottomrule
\end{tabular}
\end{table}


\section{Evaluation Metrics}
\label{sec:metric_details}

We evaluate workflow performance at two stages: \emph{retrieval} (candidates returned by the search backend) and \emph{selection} (papers retained after relevance assessment). Let $\mathcal{G}$ denote the ground-truth paper set, $\mathcal{R}_t$ the papers retrieved at iteration $t$, and $\mathcal{S}_t$ the papers selected at iteration $t$. Cumulative sets are defined as $\mathcal{R} = \bigcup_{t=1}^T \mathcal{R}_t$ and $\mathcal{S} = \bigcup_{t=1}^T \mathcal{S}_t$.

\paragraph{Retrieval-Stage Metrics.}
Retrieval precision and recall measure the quality of candidates before relevance filtering:
\begin{align}
\text{Ret.Precision} &= \frac{|\mathcal{R} \cap \mathcal{G}|}{|\mathcal{R}|}, \quad
\text{Ret.Recall} = \frac{|\mathcal{R} \cap \mathcal{G}|}{|\mathcal{G}|}.
\end{align}
Ret.F1 is the harmonic mean of Ret.Precision and Ret.Recall.

\paragraph{Selection-Stage Metrics.}
Selection precision and recall evaluate the final output:
\begin{align}
\text{Precision} &= \frac{|\mathcal{S} \cap \mathcal{G}|}{|\mathcal{S}|}, \quad
\text{Recall} = \frac{|\mathcal{S} \cap \mathcal{G}|}{|\mathcal{G}|}.
\end{align}
F1 is the harmonic mean of Precision and Recall.

\paragraph{Average Distance.}
This metric quantifies query planning effectiveness by measuring how early ground-truth papers appear in retrieval rankings. For each ground-truth paper $g \in \mathcal{G}$, let $r_g$ denote its minimum rank across all subqueries and iterations. Given a cutoff $c$ (default 100), the distance score is:
\begin{equation}
\text{dist}(g) = \max\left(1 - \frac{r_g}{c}, 0\right).
\end{equation}
Papers ranked beyond position $c$ receive zero credit. Avg.Distance averages this score:
\begin{equation}
\text{Avg.Distance} = \frac{1}{|\mathcal{G}|} \sum_{g \in \mathcal{G}} \text{dist}(g).
\end{equation}
Higher values indicate that generated subqueries successfully surface ground-truth papers near the top of retrieval results.

\paragraph{Ground-Truth Discard Rate.}
This diagnostic metric measures relevance assessment errors---ground-truth papers that were retrieved but subsequently discarded:
\begin{equation}
\text{GT Discard Rate} = \frac{|(\mathcal{R} \cap \mathcal{G}) - (\mathcal{S} \cap \mathcal{G})|}{|\mathcal{R} \cap \mathcal{G}|}.
\end{equation}
Lower values indicate better retention of relevant papers during assessment.

\paragraph{Per-Iteration Metrics.}
For iteration-level analysis, we compute cumulative metrics at each step. Let $\mathcal{R}^{(t)} = \bigcup_{i=1}^t \mathcal{R}_i$ and $\mathcal{S}^{(t)} = \bigcup_{i=1}^t \mathcal{S}_i$ denote cumulative sets through iteration $t$. Per-iteration metrics substitute these sets in the formulas above, enabling analysis of convergence patterns and marginal gains.


\section{Implementation Details}
\label{sec:workflow_details}

\subsection{Query Planning}
\label{sec:planner_details}

The query planning stage generates subqueries through structured reasoning over the current memory state. This stage maintains a subquery tree rooted at the original query (id=0) and validates consistency between requested and actual retrieval counts across iterations.

\paragraph{Subquery Generation Principles.}
Seven principles guide the planning process: (1)~preserve key concepts from the original query; (2)~avoid domain-specific terms unless explicitly relevant; (3)~employ synonyms for comprehensive coverage; (4)~vary phrasing to maximize diversity; (5)~prefer general expressions over narrow descriptions; (6)~include standard technical terminology; (7)~balance precision and recall.

\paragraph{Subquery Operations.}
Three operations govern subquery generation:
\begin{itemize}[leftmargin=*,nosep]
\item \textsc{Continue}: Request additional results for an existing subquery via pagination.
\item \textsc{Derive}: Create a specialized subquery that narrows the search scope from an existing node.
\item \textsc{Expand}: Create a parallel subquery exploring orthogonal aspects at the same granularity.
\end{itemize}

\paragraph{Memory Components.}
Two memory structures are maintained: the \emph{experience buffer} provides cumulative long-term memory across iterations; the \emph{checklist} specifies concrete retrieval criteria for each subquery to guide relevance assessment.

\paragraph{Output Schema.}
The query planning stage outputs structured JSON containing subqueries (with link type, source id, text, and target count), a checklist, an experience replay summary, and a completion flag.

\subsection{Tool Invocation}
\label{sec:tool_details}

The tool invocation stage executes retrieval calls against the static paper corpus $\mathcal{D}$. We implement two retrieval methods:

\paragraph{Dense Vector Retrieval.}
Papers are embedded using Qwen3-Embedding-0.6B on concatenated title and abstract text. Embeddings are stored in a Qdrant vector database. Given a subquery, we compute its embedding and retrieve candidates by cosine similarity. Date filtering ensures candidates satisfy temporal constraints.

\paragraph{Sparse Retrieval.}
We build a BM25 index over tokenized title and abstract text. Queries are tokenized identically, and candidates are ranked by BM25 scores. This method excels at matching specific technical terms that may be underweighted in dense representations.

Both methods support pagination via \textsc{Continue} operations. The static corpus ensures identical calls yield identical results across runs.

\subsection{Relevance Assessment}
\label{sec:assessor_details}

Relevance assessment evaluates candidates against the research objective. This stage prioritizes: direct topical alignment, clear methodological contributions, recent or seminal works, strong empirical evidence, and complete abstracts. Survey papers are down-ranked unless explicitly requested.

\paragraph{Assessment Modes.}
Two modes are implemented:

\emph{Abstract-only} mode classifies papers as selected or discarded based on titles and abstracts. This mode prioritizes efficiency.

\emph{Adaptive browsing} mode permits an ``uncertain'' classification for ambiguous cases. Papers marked \emph{to\_browse} undergo full-text extraction, after which the relevance assessment stage re-evaluates with augmented information.

\paragraph{Feedback.}
The assessment produces structured feedback: topics covered, synopsis of selected and discarded papers, and suggestions for query refinement. This propagates to query planning in subsequent iterations.

\subsection{Browser Module}
\label{sec:browser_details}

The browser module examines full-text content under adaptive browsing mode. Given a paper and an extraction goal, it fetches HTML content from ar5iv, parses hierarchical sections with mathematical notation preservation, and extracts targeted information.

The ar5iv parser extracts title, authors, abstract, and hierarchical sections. Stop-word sections (references, acknowledgments) are filtered. Mathematical expressions are preserved from \texttt{ltx\_Math} elements. The extraction produces: relevant section names, verbatim excerpts, and a synthesis of at most three sentences.


\section{Benchmark Construction}
\label{sec:dataset_details}

\paragraph{Source Processing.}
From PaSa~\citep{he2025pasa}, we extract AutoScholar (citation-derived queries) and RealScholar (human-curated queries) test sets. From LitSearch~\citep{ajith2024litsearch}, we map Semantic Scholar identifiers to arXiv IDs using external ID fields and regex extraction from PDF URLs.

\paragraph{Corpus Construction.}
Papers are merged by arXiv identifier. We query the arXiv API for complete abstracts, standardized dates, author lists, and category labels. Papers with empty abstracts, invalid identifiers, or dates outside 1990--2024 are excluded. The final corpus contains 570K papers spanning computer science, physics, and mathematics.

\paragraph{Index Construction.}
For dense retrieval, we embed papers with Qwen3-Embedding-0.6B and store vectors in Qdrant. For sparse retrieval, we build a BM25 index over tokenized text. Both indices are static, ensuring reproducibility.

\paragraph{Subset Selection.}
Test-Fast samples 200 queries with proportional source representation and ground-truth diversity. Test-Hard selects 100 challenging queries with larger ground-truth sets and cross-domain requirements.


\section{Prompt Templates}
\label{sec:prompts}

This section presents the prompt templates for each stage of our deep research workflow. All prompts adopt a structured XML-style input format and require JSON-formatted outputs for robust parsing.

\subsection{Query Planning Prompt}

This stage receives the subquery tree $\mathcal{M}_{t-1}$, experience buffer $\mathcal{B}_{t-1}$, and feedback $\mathcal{O}_{t-1}$ to produce plan $\mathcal{P}_t$ (cf.\ Section~\ref{sec:workflow}). The prompt encodes three operations---\textsc{Continue}, \textsc{Derive}, and \textsc{Expand}---along with subquery generation strategies. Table~\ref{tab:query_planning_reference} details the prompt components; Figure~\ref{fig:planner_prompt} presents the template; Figure~\ref{fig:planner_output_example} shows a representative output.

\begin{table*}[htbp]
\centering
\small
\caption{Query Planning prompt components: strategies, linking guidance, and field reference.}
\label{tab:query_planning_reference}
\renewcommand{\arraystretch}{1.15}
\begin{tabular}{@{}ll@{\hspace{8pt}}p{11.8cm}@{}}
\toprule
\multicolumn{3}{l}{\textbf{Panel A: Strategies} (Subquery Generation Principles)} \\
\midrule
& S1 & Preserve all key points from the user's original query. \\
& S2 & Avoid domain-specific keywords unless explicitly requested or strongly domain-tied. \\
& S3 & Use synonyms and near-synonyms to capture different expressions of the same concept. \\
& S4 & Avoid repetitive sentence structures; vary phrasing to improve diversity. \\
& S5 & Prefer general, widely-used expressions over narrow descriptions. \\
& S6 & Include professional and technical terms commonly used in the field. \\
& S7 & Combine multiple strategies to generate queries that are both precise and diverse. \\
\midrule
\multicolumn{3}{l}{\textbf{Panel B: Linking Guidance} (Subquery Operations)} \\
\midrule
& \textsc{Continue} & Request additional results for an existing subquery via pagination. \\
&                   & \textit{Required}: \texttt{source\_id}, \texttt{target\_k}. \\
& \textsc{Derive}   & Create a specialized subquery that deepens exploration from an existing node. \\
&                   & \textit{Required}: \texttt{source\_id}, \texttt{text}, \texttt{target\_k}. \\
& \textsc{Expand}   & Create a parallel subquery exploring alternative aspects at the same granularity. \\
&                   & \textit{Required}: \texttt{source\_id}, \texttt{text}, \texttt{target\_k}. \\
& \textit{Rules}    & The subquery space is a tree rooted at id=0. Every \texttt{source\_id} must reference a node in \texttt{<all\_subqueries>}. Root (id=0) is valid for \textsc{Derive}/\textsc{Expand} but not for \textsc{Continue}. \\
\midrule
\multicolumn{3}{l}{\textbf{Panel C: Field Reference} (Input Fields)} \\
\midrule
& \texttt{all\_subqueries} & Each line: id, source (parent id), link type, iteration, text. \\
& \texttt{last\_iteration\_state} & Per subquery: id, text, target\_k, retrieved, selected, overview. \\
& \texttt{previous\_iteration\_state} & Earlier iteration states, same structure as above. \\
& \texttt{last\_checklist} & Concrete retrieval criteria from the previous round. \\
& \texttt{last\_experience\_replay} & Cumulative long-term memory across iterations. \\
\bottomrule
\end{tabular}
\end{table*}

\begin{figure*}[htbp]
\begin{tcolorbox}[colback=gray!5, colframe=black!75, title=\textbf{Prompt: Query Planning}, fontupper=\small]
\textbf{[System]}\\
You are the Query Planning module in a Deep Research workflow. Analyze the user query and the research memory to propose subqueries and set target\_k (papers per subquery). The subquery space is a tree rooted at id=0 (the original user query). You may choose any existing node (including id=0) as the source for DERIVE/EXPAND, and any non-root existing node for CONTINUE. Validate the last\_iteration\_state against the last\_checklist to ensure consistency (e.g., whether requested target\_k and actually retrieved/selected counts align). The Relevance Assessment module provides overview for all subqueries from all previous rounds. The most recent overviews can be found in last\_iteration\_state, while earlier ones are stored in previous\_iteration\_state. Use experience\_replay as the cumulative long-term memory: integrate last\_experience\_replay, insights from last\_iteration\_state, and your current planning rationale; emphasize what changed since the previous iteration. Maintain a concise checklist of concrete retrieval criteria per subquery (what to retrieve next: required methods, tasks, datasets, time ranges, include/exclude signals) for the Relevance Assessment module and the next Query Planning iteration. Each retrieval call returns exactly one page with up to \{max\_results\_per\_request\} results. To access additional pages, issue a CONTINUE operation. The root query (id=0) cannot be continued. Think step-by-step and keep outputs structured for machine-readability.\\[0.5em]
\textbf{[User]}\\
\texttt{<user\_query>}\{user\_query\}\texttt{</user\_query>}\\
\texttt{<current\_iteration>}\{current\_iteration\}\texttt{</current\_iteration>}\\
\texttt{<last\_iteration\_state>}\{last\_iteration\_state\}\texttt{</last\_iteration\_state>}\\
\texttt{<previous\_iteration\_state>}\{previous\_iteration\_state\}\texttt{</previous\_iteration\_state>}\\
\texttt{<last\_checklist>}\{last\_checklist\}\texttt{</last\_checklist>}\\
\texttt{<last\_experience\_replay>}\{last\_experience\_replay\}\texttt{</last\_experience\_replay>}\\
\texttt{<all\_subqueries>}\{all\_subqueries\}\texttt{</all\_subqueries>}\\
\texttt{<strategies>}(See Table~\ref{tab:query_planning_reference})\texttt{</strategies>}\\
\texttt{<linking\_guidance>}(See Table~\ref{tab:query_planning_reference})\texttt{</linking\_guidance>}\\
\texttt{<field\_reference>}(See Table~\ref{tab:query_planning_reference})\texttt{</field\_reference>}\\[0.3em]
\texttt{<instructions>}\\
1) Plan \& Validate: Briefly reflect on research gaps and progress. Validate \texttt{<last\_iteration\_state>} against \texttt{<last\_checklist>} (note any mismatches between requested target\_k vs. actual retrieval counts).\\
2) Subqueries Strategy: CONTINUE, DERIVE, EXPAND.\\
3) Subqueries: Propose 3--6 keyword-style subqueries with integer target\_k each. For each item, choose source\_id from \texttt{<all\_subqueries>} (including 0 for DERIVE/EXPAND) and set link\_type.\\
4) Checklist: Write concrete retrieval criteria tailored to the current subqueries (what to retrieve/include/exclude; methods, tasks, datasets, metrics, temporal filters).\\
5) Experience Replay: Update long-term memory by integrating \texttt{<last\_experience\_replay>} with insights from \texttt{<last\_iteration\_state>} and current rationale. Preserve historical key points and explicitly note what changed vs.\ the previous iteration.\\
6) Completion: Set is\_complete=true ONLY if existing retrieved information is sufficient to comprehensively answer the user query.\\
\texttt{</instructions>}\\[0.3em]
\texttt{<output\_format>}\\
Provide JSON only inside \texttt{<planner\_output>} tags:\\
\texttt{<planner\_output>}\\
\texttt{\{"subqueries": [\{"link\_type": "...", "source\_id": xx, "text": "...", "target\_k": xx\}, ...],}\\
\texttt{ "checklist": "...", "experience\_replay": "...", "is\_complete": false\}}\\
\texttt{</planner\_output>}\\
\texttt{</output\_format>}
\end{tcolorbox}
\caption{Query Planning prompt template with full history and state validation.}
\label{fig:planner_prompt}
\end{figure*}

\begin{figure*}[htp]
\centering
\small
\begin{tcolorbox}[colback=gray!5, colframe=gray!80, boxrule=0.5pt, arc=2mm]
\textbf{Output Example:}
\begin{verbatim}
{
  "subqueries": [
    {"link_type": "continue", "source_id": 3, "target_k": 10},
    {"link_type": "derive", "source_id": 4, 
     "text": "self-normalized IPS SNIPS Wang 2017", "target_k": 10},
    {"link_type": "expand", "source_id": 0, 
     "text": "IPS variance reduction bias correction", "target_k": 10}
  ],
  "checklist": "Prioritize IPS/SNIPS for selection bias. Target foundational 
     SNIPS paper (Wang et al. 2017). Include theoretical analyses...",
  "experience_replay": "Iter 1: IPS/SNIPS concentrated in counterfactual 
     learning. Foundational paper missing. Iter 2 derives targeted queries...",
  "is_complete": false
}
\end{verbatim}
\end{tcolorbox}
\caption{Output example from Query Planning.}
\label{fig:planner_output_example}
\end{figure*}

\subsection{Relevance Assessment Prompts}

This stage evaluates candidates $\{\mathcal{C}_i\}$ against plan $\mathcal{P}_t$ and outputs selected papers $\mathcal{S}_t$ with feedback $\mathcal{O}_t$ (cf.\ Section~\ref{sec:workflow}). We implement two modes.

\subsubsection{Abstract-Only Mode}

Candidates are classified based on metadata alone. Figure~\ref{fig:selector_prompt_abstract} presents the template; Figure~\ref{fig:selector_output_abstract} shows a representative output.

\begin{figure*}[htbp]
\begin{tcolorbox}[colback=gray!5, colframe=black!75, title=\textbf{Prompt: Relevance Assessment (Abstract-only)}, fontupper=\small]
\textbf{[System]}\\
You are the Relevance Assessment module. Given a subquery, the Query Planning checklist, the assessment recipe, and candidate papers with retriever scores, decide which papers to maintain. Be strict; maintain only high-quality, directly relevant papers. Output decisions in JSON. Additionally, provide a structured overview of the retrieval and selection results to help the Query Planning module adjust future subqueries.\\[0.5em]
\textbf{[User]}\\
\texttt{<original\_query>}\{user\_query\}\texttt{</original\_query>}\\
\texttt{<sub\_query>}\{sub\_query\}\texttt{</sub\_query>}\\
\texttt{<planner\_checklist>}\{planner\_checklist\}\texttt{</planner\_checklist>}\\
\texttt{<selector\_recipe>}\{selector\_recipe\}\texttt{</selector\_recipe>}\\
\texttt{<candidates>}\{candidates\}\texttt{</candidates>}\\[0.3em]
\texttt{<instructions>}\\
Return JSON inside \texttt{<selector\_output>...</selector\_output>}. Fields:\\
- selected: array of paper\_id kept\\
- reasons: mapping paper\_id $\rightarrow$ short reason\\
- overview: a single string summarizing the result, including:\\
\hspace*{1em}1) retrieved\_topics: what topics the retrieved papers cover\\
\hspace*{1em}2) relevant\_summary: what the selected papers discuss\\
\hspace*{1em}3) irrelevant\_summary: what discarded papers discuss and why irrelevant\\
\hspace*{1em}4) adjustment\_suggestions: optional suggestions for refining the subquery\\
\texttt{</instructions>}\\[0.3em]
\texttt{<output\_format>}\\
\texttt{<selector\_output>}\\
\texttt{\{"selected": [], "reasons": \{\}, "overview": "..."\}}\\
\texttt{</selector\_output>}\\
\texttt{</output\_format>}
\end{tcolorbox}
\caption{Relevance Assessment prompt template (Abstract-only mode).}
\label{fig:selector_prompt_abstract}
\end{figure*}

\begin{figure*}[htp]
\centering
\small
\begin{tcolorbox}[colback=gray!5, colframe=gray!80, boxrule=0.5pt, arc=2mm]
\textbf{Output Example:}
\begin{verbatim}
{
  "selected": ["1503.02045"],
  "reasons": {
    "1503.02045": "Addresses estimation after parameter selection, 
       discussing selection bias and corrective estimators."
  },
  "overview": "1.retrieved_topics: Wireless comm, numerical methods, 
     covariance selection. 2.relevant_summary: Selected paper tackles 
     selection bias in parameter estimation. 3.irrelevant_summary: Most 
     papers discuss 'selection' in unrelated contexts. 4.adjustment: 
     Use 'Inverse Propensity Score weighting selection bias'."
}
\end{verbatim}
\end{tcolorbox}
\caption{Output example from Relevance Assessment (Abstract-only).}
\label{fig:selector_output_abstract}
\end{figure*}

\subsubsection{Adaptive Browsing Mode}

An intermediate \texttt{to\_browse} label triggers full-text extraction before final classification. Figure~\ref{fig:selector_prompt_adaptive} presents the template; Figure~\ref{fig:selector_output_adaptive} shows a representative output.

\begin{figure*}[htbp]
\begin{tcolorbox}[colback=gray!5, colframe=black!75, title=\textbf{Prompt: Relevance Assessment (Adaptive Browsing)}, fontupper=\small]
\textbf{[System]}\\
You are the Relevance Assessment module. Given a subquery, the Query Planning checklist, the assessment recipe, and candidate papers (meta-info \textbf{and/or browser evidence}), decide which papers to select, discard, or browse further. If information is insufficient, delegate to the Browser. Be strict on selection; be precise on browsing goals. Output decisions in JSON. Additionally, provide a structured overview to help the Query Planning module adjust future subqueries.\\[0.5em]
\textbf{[User]}\\
\texttt{<original\_query>}\{user\_query\}\texttt{</original\_query>}\\
\texttt{<sub\_query>}\{sub\_query\}\texttt{</sub\_query>}\\
\texttt{<planner\_checklist>}\{planner\_checklist\}\texttt{</planner\_checklist>}\\
\texttt{<selector\_recipe>}\{selector\_recipe\}\texttt{</selector\_recipe>}\\
\texttt{<context>}\texttt{<candidates>}\{candidates\}\texttt{</candidates>}\texttt{</context>}\\[0.3em]
\texttt{<instructions>}\\
1. \textbf{Finalize Decisions (with Browser Evidence)}: For candidates that include \texttt{browser\_summary} data, make a definitive decision: include them in \texttt{selected} or \texttt{discarded}. Do not put these candidates into \texttt{to\_browse}.\\
2. \textbf{Handle Meta-info Only Candidates}: For candidates with only meta-info (title, abstract...): if relevance is clear, add to \texttt{selected} or \texttt{discarded}; if relevance is plausible but uncertain, add to \texttt{to\_browse} with a specific extraction goal.\\
3. \textbf{Generate Overview}: Generate an overview based strictly on the current batch of candidates and decisions. Summarize insights gained from both selected papers and browser results to help the Query Planning module refine future subqueries.\\
4. \textbf{JSON Requirement}: Output the result strictly in JSON format inside \texttt{<selector\_output>}.\\[0.3em]
Fields (ensure all paper\_id references use arxiv\_id format):\\
- selected: array of paper\_id kept as high-quality, relevant evidence\\
- discarded: array of paper\_id excluded from further consideration\\
- to\_browse: mapping paper\_id $\rightarrow$ specific extraction goal for the Browser\\
- reasons: mapping paper\_id $\rightarrow$ short reason for the current decision\\
- overview: a single string (retrieved\_topics, relevant\_summary, irrelevant\_summary, adjustment\_suggestions)\\
\texttt{</instructions>}\\[0.3em]
\texttt{<output\_format>}\\
\texttt{<selector\_output>}\\
\texttt{\{"selected": [], "discarded": [], "to\_browse": \{\}, "reasons": \{\}, "overview": "..."\}}\\
\texttt{</selector\_output>}\\
\texttt{</output\_format>}
\end{tcolorbox}
\caption{Relevance Assessment prompt template (Adaptive Browsing mode).}
\label{fig:selector_prompt_adaptive}
\end{figure*}

\begin{figure*}[htp]
\centering
\small
\begin{tcolorbox}[colback=gray!5, colframe=gray!80, boxrule=0.5pt, arc=2mm]
\textbf{Output Example:}
\begin{verbatim}
{
  "selected": ["1003.5956"],
  "discarded": ["1504.06937", "1310.1404", "1407.2806"],
  "to_browse": {},
  "reasons": {
    "1003.5956": "Seminal paper on unbiased offline evaluation, 
       connects to IPS/SNIPS correction for selection bias.",
    "1504.06937": "Constrained bandits regret. No IPS/SNIPS."
  },
  "overview": "1.retrieved_topics: Contextual bandits, cold-start. 
     2.relevant_summary: One paper proposes unbiased offline evaluation. 
     3.irrelevant_summary: Nine papers focus on online bandit algorithms. 
     4.adjustment: Include 'unbiased offline evaluation'."
}
\end{verbatim}
\end{tcolorbox}
\caption{Output example from Relevance Assessment (Adaptive Browsing).}
\label{fig:selector_output_adaptive}
\end{figure*}

\subsection{Browser Extraction Prompt}

This module extracts targeted evidence from full-text when invoked by adaptive browsing. Figure~\ref{fig:browser_prompt} presents the template; Figure~\ref{fig:browser_output_example} shows a representative output.

\begin{figure*}[htbp]
\begin{tcolorbox}[colback=gray!5, colframe=black!75, title=\textbf{Prompt: Browser Extraction}, fontupper=\small]
\textbf{[System]}\\
You are the Paper Extraction module. Your goal is to extract the minimum viable evidence from a paper to satisfy a specific research goal. Focus on precision over volume. Capture technical verbatim data accurately but eliminate tangential context.\\[0.5em]
\textbf{[User]}\\
\texttt{<paper\_content>}\{full\_text\}\texttt{</paper\_content>}\\
\texttt{<user\_goal>}\{task\}\texttt{</user\_goal>}\\[0.3em]
\texttt{<instructions>}\\
Return JSON inside \texttt{<extractor\_output>}. All fields must be Strings:\\
- rational: List only the specific section names relevant to the goal.\\
- evidence: Extract concise, verbatim snippets that directly support the goal. Use ellipses (...) to skip irrelevant filler.\\
- summary: A max 3-sentence synthesis stating what the paper provides regarding the goal.\\
\texttt{</instructions>}\\[0.3em]
\texttt{<output\_format>}\\
\texttt{<extractor\_output>}\\
\texttt{\{"rational": "...", "evidence": "...", "summary": "..."\}}\\
\texttt{</extractor\_output>}\\
\texttt{</output\_format>}
\end{tcolorbox}
\caption{Browser Extraction prompt template.}
\label{fig:browser_prompt}
\end{figure*}

\begin{figure*}[htp]
\centering
\small
\begin{tcolorbox}[colback=gray!5, colframe=gray!80, boxrule=0.5pt, arc=2mm]
\textbf{Output Example:}
\begin{verbatim}
{
  "rational": "Section 2: Related Work, Section 7: References",
  "evidence": "Point-based rendering traces back to Levoy [1985]... 
     Zwicker et al. [2001] introduced surface splatting... 
     Yifan et al. [2019] proposed differentiable surface splatting...",
  "summary": "Survey traces Gaussian-based rendering lineage from 
     Zwicker (2001) through Yifan (2019) to 3D Gaussian Splatting. 
     Confirms historical connection between EWA and differentiable 
     point-based graphics."
}
\end{verbatim}
\end{tcolorbox}
\caption{Output example from Browser Extraction.}
\label{fig:browser_output_example}
\end{figure*}


\end{document}